\documentclass[letterpaper]{article} 
\PassOptionsToPackage{table}{xcolor} 
\usepackage[preprint]{aaai2027}  
\usepackage[hyphens]{url}  
\usepackage{graphicx} 
\usepackage{natbib}  
\usepackage{caption} 
\usepackage{booktabs}
\usepackage{colortbl}
\usepackage{amsmath, amssymb, amsthm}
\usepackage{multirow}
\usepackage{microtype}

\title{CVSD-Reg: Cross-Modal Visual Semantic Prior\\
Distillation for Robust LiDAR Registration}

\author{
    Eunsoo Im,
    Gyeonggwan Lee,
    Seunghwan Hong,
    and Junghun Suh
}
\affiliations{
    Kakao Mobility\\
    \{soo.hd, gandan.lee, logan.sh, jude.suh\}@kakaomobility.com
}

\begin{document}
\maketitle

\begin{abstract}

Learning-based global point cloud registration has achieved remarkable progress,
yet its reliance on geometric representations makes existing methods sensitive to variations in point density,
scan pattern, viewpoint, and sensor characteristics.
We propose CVSD-Reg, a robust global LiDAR registration framework that distills visual semantic priors from a vision foundation model into LiDAR representations.
In Stage 1, a Point Transformer V3 student learns from a frozen DINOv2 teacher through contrastive distillation and spherical-manifold alignment,
which encourages pointwise angular alignment to the projected teacher embedding.
Self-supervised InfoNCE consistency and soft $\mathrm{SE}(3)$ invariance further encourage descriptors that are robust under rigid transformations of the same scan.
In Stage 2, the distilled representation is adapted to registration through correspondence learning,
density-aware point-dropout augmentation, and end-to-end pose optimization.
With a single checkpoint, CVSD-Reg generalizes to both single-sensor and zero-shot cross-sensor scenarios without sensor-specific adaptation and remains entirely camera-free at inference.
On KITTI, nuScenes, and HeLiPR, CVSD-Reg achieves strict success rates (SR@0.5\,m/$1^\circ$) of 97.7$\%$,
99.0$\%$, and 99.3$\%$, respectively, including 97.3$\%$ on sparse 16-beam Velodyne scans.
Relative to BUFFER-X, the strongest overall geometric baseline on HeLiPR, it gains up to $44.0$ percentage points on 16-beam scans ($+37.3$ over the best baseline on that sensor), without requiring camera inputs or post-hoc ICP refinement.
Project page: \url{https://ies0411.github.io/CVSD_prj/}.
\end{abstract}

\section{Introduction}

Point cloud registration estimates a rigid transformation $\mathbf{T}\in\mathrm{SE}(3)$ 
that aligns a source scan $\mathcal{P}_{q}$ with a target scan $\mathcal{P}_{t}$ 
and is fundamental to mapping, localization, and multi-session map updates.
Although recent learning-based methods have substantially advanced global registration, 
their descriptors and correspondences remain sensitive to geometric variations 
arising from changes in point density, scan pattern, viewpoint, and
sensor characteristics.
Both classical descriptors such as FPFH~\cite{rusu2009fast} and learned
geometric matchers~\cite{geotransformer,cast} can therefore degrade under
severe sparsity, low overlap, or out-of-distribution sampling patterns.
This problem is particularly pronounced across heterogeneous LiDAR systems, 
whose different beam configurations and sensing principles
induce substantial shifts in observed geometry, 
motivating complementary priors that remain informative 
when geometric evidence is unreliable.


Recent registration methods have sought to improve robustness to such geometric variations 
through adaptive voxelization and geometric normalization~\cite{seo2025buffer}
or flow-matching-based registration~\cite{rap}.
Nevertheless, their correspondences remain predominantly geometry-driven and
can become unreliable when structural evidence is sparse or ambiguous.
Vision foundation models such as DINOv2~\cite{dinov2} 
provide complementary semantic representations
that are comparatively stable across viewpoints
and can disambiguate geometrically similar structures.
Prior works, however, either require visual features at inference time~\cite{vfmreg} 
or distill them mainly for semantic understanding~\cite{slidr,scalr}, 
leaving their transfer to camera-free LiDAR descriptors 
for correspondence estimation and pose recovery largely unexplored.


To bridge this gap, we propose CVSD-Reg, a robust global LiDAR registration framework
that learns 3D representations robust to geometric variations by distilling visual semantic priors.
Our key idea is to decouple semantic representation learning
from registration-specific adaptation in a two-stage training scheme.
In Stage 1, visual semantic priors encoded by a frozen vision foundation model
are transferred to a LiDAR backbone through cross-modal contrastive distillation.
To retain angular organization during distillation,
teacher and student descriptors are further aligned on the unit hypersphere.
This encourages pointwise angular alignment in the projected teacher space,
while consistency learning across rigidly transformed views
promotes robustness to rigid transformations of the same point set.
In Stage 2, the pretrained representation is adapted to pairwise registration
by jointly learning cross-scan correspondences and relative pose estimation.
Confidence-weighted matches are integrated with a differentiable Kabsch solver,
allowing pose-level supervision to directly refine both the descriptor and correspondence networks,
while density-aware point dropout improves robustness to variations in scan sparsity.


CVSD-Reg uses a single checkpoint and a unified inference configuration
across conventional single-sensor benchmarks and zero-shot cross-sensor settings.
It achieves a strict success rate (SR@0.5\,m/$1^\circ$) of 97.7\% on KITTI, 99.0\% on nuScenes, and 99.3\% on HeLiPR,
including 97.3\% on sparse 16-beam Velodyne scans.  
These results indicate that visual semantic prior distillation improves registration robustness
to geometric distribution shifts without compromising standard single-sensor performance.


In summary, our main contributions are three-fold:
\begin{enumerate}
\item
  We present CVSD-Reg, a global LiDAR registration framework
  that distills visual semantic priors from a frozen vision foundation model into a LiDAR backbone,
  enabling camera-free inference.

\item
  We introduce a registration-oriented two-stage learning strategy
  that couples hyperspherical cross-modal distillation in Stage 1
  with density-robust correspondence and end-to-end pose supervision in Stage 2.

\item
  A single CVSD-Reg checkpoint achieves strong single-sensor accuracy on KITTI and nuScenes
  and state-of-the-art zero-shot cross-sensor robustness on HeLiPR,
  gaining up to $44.0$ percentage points over BUFFER-X, the strongest overall geometric baseline,
  on sparse 16-beam scans ($+37.3$ over the best baseline on that sensor).

\end{enumerate}

\begin{figure*}[t]
\centering
\includegraphics[width=\linewidth]{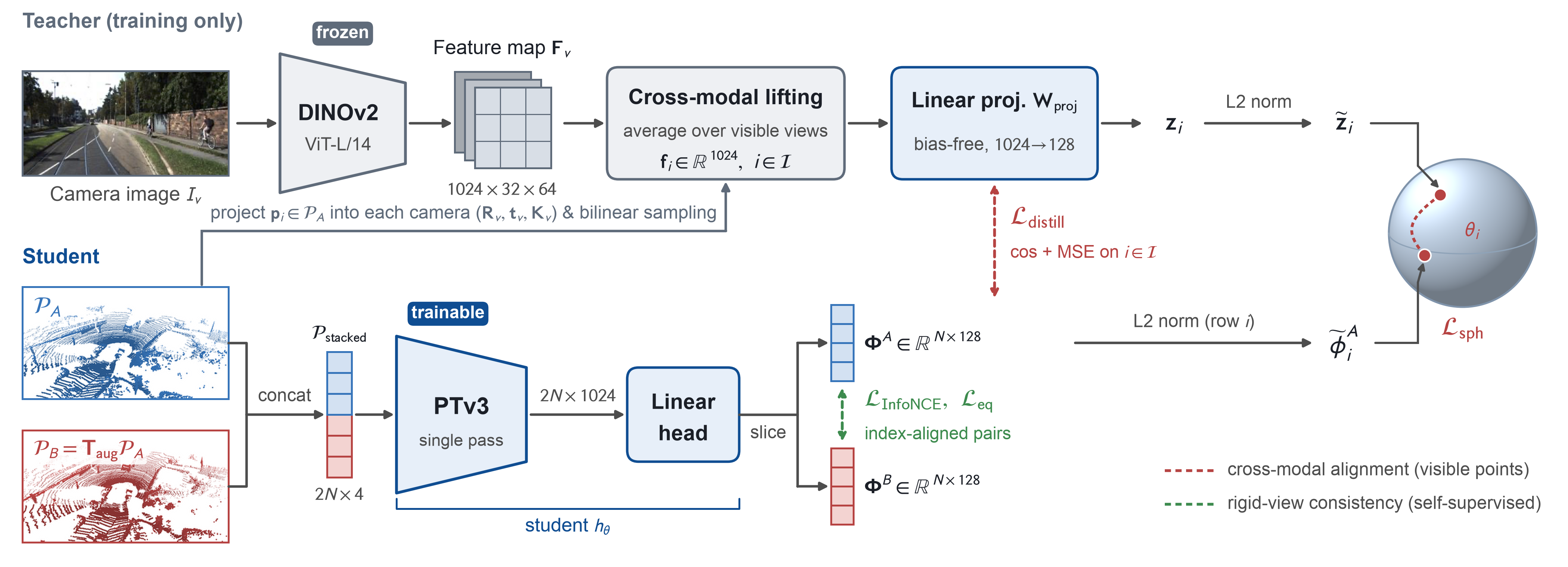}
\caption{Overview of Stage-1 cross-modal distillation framework.
(Top) Teacher branch: Dense 2D DINOv2 features are lifted onto visible 3D LiDAR points via projective geometry and bilinear interpolation.
(Bottom) Student branch: A trainable PTv3 backbone processes original ($\mathcal{P}_{A}$) and augmented ($\mathcal{P}_{B}$) scans in a single stacked pseudo-batch pass.
The representation is optimized jointly via cross-modal alignment ($\mathcal{L}_{\text{distill}}, \mathcal{L}_{\text{sph}}$) within the visibility mask $\mathcal{I}$ and self-supervised rigid-view consistency ($\mathcal{L}_{\text{InfoNCE}}, \mathcal{L}_{\text{eq}}$).}
\label{fig:stage1}
\end{figure*}

\section{Related Work}
\label{sec:related}

\paragraph{Cross-modal distillation}
SLidR~\cite{slidr} (MoCo~v2 ResNet-50 teacher) and ScaLR~\cite{scalr} distill image features into LiDAR for semantic segmentation, not registration.
PointCLIP~\cite{zhang2022pointclip} uses depth-map rendering for CLIP-style recognition, and OpenScene~\cite{peng2023openscene} pursues open-vocabulary 3D understanding.
VFM-Registration~\cite{vfmreg} projects DINOv2 onto outdoor LiDAR at test time and needs cameras at deployment; DINOReg~\cite{chen2025dinoreg} combines DINOv2 with geometric cues on indoor RGB-D and KITTI, but does not target zero-shot cross-sensor outdoor LiDAR registration.
BYOC~\cite{banani2021byoc} learns RGB-D geometric features without a foundation teacher, and SuperFlow~\cite{xu2024superflow} targets cross-modal pretraining for segmentation; neither addresses zero-shot cross-sensor outdoor LiDAR registration.
CVSD-Reg distills once for registration-oriented descriptors and runs camera-free at inference on heterogeneous outdoor LiDAR.

\paragraph{LiDAR registration}
Learning-based matchers such as GeoTransformer~\cite{geotransformer}, CAST~\cite{cast}, PARE-Net~\cite{parenet}, and UGP~\cite{ugp} attain strong in-distribution accuracy,
but their descriptors are tied to the multi-beam geometries seen in training and degrade under unseen sparsity or scan patterns.
BUFFER-X~\cite{seo2025buffer} mitigates some of this gap with adaptive voxelization and patch-wise scale normalization for zero-shot scene transfer,
while RAP~\cite{rap} reformulates registration as conditional flow matching with an alternating-attention velocity field.
Learning-free pipelines such as MAC~\cite{mac}, a compatibility-based estimator, and KISS-Matcher~\cite{lim2025kiss}, which improves scalability via Faster-PFH descriptors and linear-complexity graph pruning, still depend on handcrafted geometric cues.
Across both families, purely geometric evidence becomes unreliable when beam density collapses or sensor topology shifts out of distribution.
As a complementary approach, CVSD-Reg retains a standard 3D backbone and injects vision-derived semantic anchors during training to stabilize matching under these shifts.

\paragraph{Feature geometry}
Registration quality depends on how descriptors are organized in feature space.
Hard structural equivariance, as in PARE-Net~\cite{parenet}, $\mathrm{SE}(3)$-Transformers~\cite{fuchs2020se3}, and Vector Neurons~\cite{deng2021vn}, preserves pose structure by construction.
Hyperspherical metric learning such as ArcFace~\cite{deng2019arcface} and SphereFace~\cite{liu2017sphereface} instead sharpens angular separability on the unit sphere.
Rather than imposing hard equivariant architectures, CVSD-Reg adopts a soft alternative.
Student features are aligned to the teacher on $\mathbb{S}^{d-1}$ via geodesic distance with an angular margin,
and rigid-view consistency is encouraged with a soft $\mathrm{SE}(3)$ invariance objective,
which avoids the distortion that flat Euclidean matching can induce.

\section{Overview}

CVSD-Reg adopts a two-stage training framework 
that separates visual semantic prior learning from registration-specific adaptation. 
Stage 1 distills semantic priors from a frozen DINOv2~\cite{dinov2} teacher 
into a Point Transformer V3 (PTv3)~\cite{wu2024point} LiDAR backbone using synchronized image--LiDAR data (Figure~\ref{fig:stage1}). 
Hyperspherical teacher--student alignment and 
consistency learning across rigidly transformed LiDAR views 
encourage semantically grounded descriptors with rigid-view consistency. 
In Stage 2, the vision teacher is removed, 
and the pretrained LiDAR backbone is fine-tuned on unposed LiDAR pairs 
through correspondence and pose learning under varying scan densities, 
as detailed in Figure~\ref{fig:stage2}. 
This decoupling confines visual supervision to pretraining 
and yields a camera-free LiDAR-only registration pipeline at inference.

\section{Stage-1: Cross-Modal Distillation Pretraining}
\label{sec:stage1}

Given a LiDAR scan $\mathcal{P}_A = \{\mathbf{p}_i\}_{i=1}^{N}$ with coordinates $\mathbf{p}_i \in \mathbb{R}^3$ and per-point intensity,
and $V$ synchronized camera images $\{I_v\}_{v=1}^V$, 
Stage 1 transfers dense visual semantic priors from the frozen DINOv2 teacher to the PTv3 student. 
The teacher branch lifts image features onto LiDAR points visible from the synchronized cameras,
providing point-wise semantic supervision. 
The student branch encodes the original scan and a rigidly augmented view. 
The representation is jointly optimized 
through cross-modal teacher--student distillation and self-supervised rigid-view consistency.
\paragraph{Visual Feature Lifting}

To construct point-wise semantic targets, 
we lift dense DINOv2 features from the image plane to the LiDAR coordinate frame.
For the $v$-th camera, 
the point $\mathbf{p}_{i}$ is transformed and projected 
onto the image plane as
\begin{equation}
\mathbf{q}_{i}^{v} =
\mathbf{R}_{v}\mathbf{p}_{i} + \mathbf{t}_{v},
\qquad \mathbf{u}_{i}^{v}
 = \pi\left(\mathbf{K}_{v}\mathbf{q}_{i}^{v}\right),
\label{eq:point_projection}
\end{equation}
where $\mathbf{R}_{v}\in\mathrm{SO}(3)$ and $\mathbf{t}_{v}\in\mathbb{R}^{3}$
define the LiDAR-to-camera transformation, 
$\mathbf{K}_{v}$ is the camera intrinsic matrix,
and $\pi(\cdot)$ denotes perspective division.

\begin{figure*}[t]
\centering
\includegraphics[width=\linewidth]{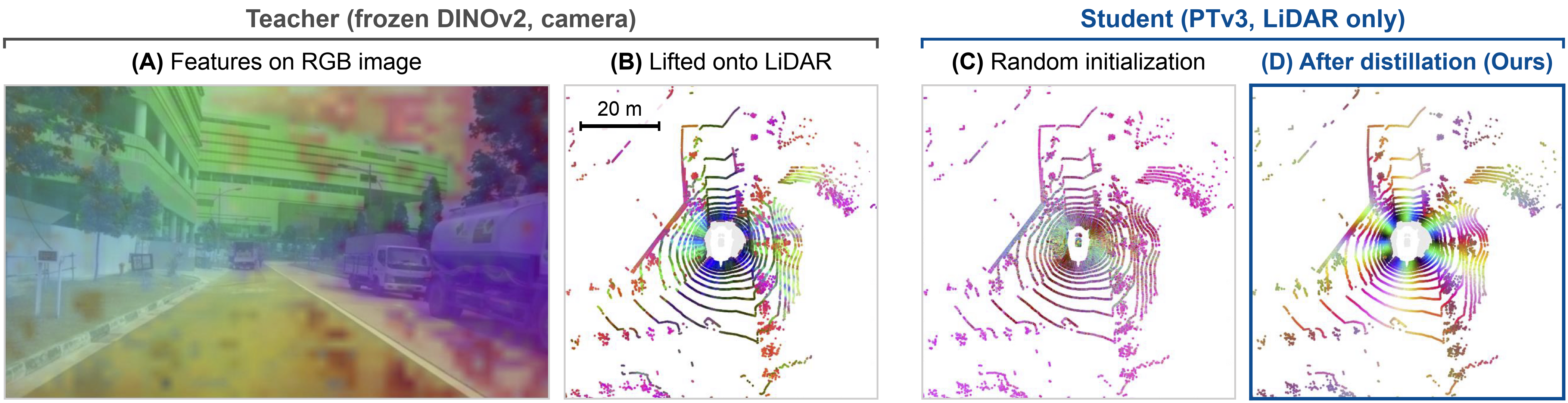}
\caption{%
    \textbf{Cross-modal distillation visualized by PCA feature coloring.}
    High-dimensional descriptors are projected to RGB via PCA, so nearby colors indicate similar feature directions.
    (A)~DINOv2 features on the RGB frame (\texttt{CAM\_FRONT}).
    (B)~The same teacher features lifted onto LiDAR.
    (C)~A randomly initialized student yields nearly uniform colors (no semantic structure).
    (D)~After distillation, the LiDAR-only student recovers a color topology closely matching the teacher in~(B).
}
\label{fig:distillation_quality}
\end{figure*}

Let $\mathcal{V}_{i}$ denote the set of cameras in which
$\mathbf{p}_{i}$ has a valid depth in $[z_{\min}, z_{\max}]$\,m and projects inside the image boundary.
For each valid view, the DINOv2 feature is bilinearly sampled at $\mathbf{u}_i^v$
as $\mathbf{F}_{v}(\mathbf{u}_{i}^{v})$.
The point-wise teacher feature is then computed by averaging these
features over all valid views:
\begin{equation}
\mathbf{f}_{i}
=
\frac{1}{\lvert\mathcal{V}_{i}\rvert}
\sum_{v\in\mathcal{V}_{i}}
\mathbf{F}_{v}\left(\mathbf{u}_{i}^{v}\right),
\label{eq:teacher_feature_lifting}
\end{equation}
Cross-modal supervision is applied only to visible points indexed by 
$\mathcal{I}=\{i\in\{1,\ldots,N\}\mid\lvert\mathcal{V}_i\rvert>0\}$.
Figure~\ref{fig:distillation_quality} qualitatively compares the lifted teacher features with the randomly initialized and distilled student representations.

\paragraph{Dual-View LiDAR Encoding}

The student branch extracts descriptors from the original scan
$\mathcal{P}_{A}$ and a rigidly augmented view
$\mathcal{P}_{B} = \mathbf{T}_{\mathrm{aug}}\mathcal{P}_{A}$,
where $\mathbf{T}_{\mathrm{aug}}\in\mathrm{SE}(3)$ is randomly sampled.
Because the transformation is applied without resampling,
the points retain index-wise correspondences across the two views.
The two views are jointly encoded with shared parameters in a single PTv3 forward pass:
\begin{equation}
\begin{aligned}
\left[\mathbf{\Phi}^{A};\mathbf{\Phi}^{B}\right]
&=
h_{\theta}\!\left(\mathcal{P}_{stacked}
\right),
&\quad
\mathbf{\Phi}^{A},\mathbf{\Phi}^{B} \in\mathbb{R}^{N \times d}.
\end{aligned}
\label{eq:dual_view_encoding}
\end{equation}
Here, $\mathcal{P}_{stacked}$ denotes the batch-wise concatenation 
of $\mathcal{P}_{A}$ and $\mathcal{P}_{B}$,
and $h_{\theta}$ is the PTv3 student.
The matrices $\mathbf{\Phi}^{A}$ and $\mathbf{\Phi}^{B}$ contain the corresponding 
$d$-dimensional point descriptors for the original and transformed views, respectively.

\paragraph{Hyperspherical Cross-Modal Distillation}

We transfer the lifted visual priors to the LiDAR representation 
by aligning the teacher and student descriptors in a common embedding space.
A bias-free linear projector $\mathbf{W}_{\mathrm{proj}}$ 
maps each point-wise teacher feature $\mathbf{f}_{i}$ 
into the $d$-dimensional student embedding space:
\begin{equation}
    \mathbf{z}_{i}
    =
    \mathbf{W}_{\mathrm{proj}}\mathbf{f}_{i}
    \in
    \mathbb{R}^{d}.
    \label{eq:teacher_projection}
\end{equation}
Here, $\mathbf{z}_{i}$ is the teacher feature projected 
into the student embedding space for point $\mathbf{p}_{i}$.
Let $\boldsymbol{\phi}_{i}^{A}\in\mathbb{R}^{d}$ denote the $i$-th row of $\mathbf{\Phi}^{A}$.
The primary distillation objective combines directional alignment with feature regression:
\begin{equation}
\label{eq:l_distill}
\begin{aligned}
\mathcal{L}_{\text{distill}} = {} & 
\frac{1}{|\mathcal{I}|} \sum_{i \in \mathcal{I}} 
\left( 1 - \left\langle \widetilde{\boldsymbol{\phi}}_i^{A}, \widetilde{\mathbf{z}}_i \right\rangle \right) \\
& + \lambda_{\mathrm{mse}}\operatorname{MSE}\!\left( \mathbf{\Phi}_{\mathcal{I}}^{A},\, \mathbf{Z}_{\mathcal{I}} \right),
\end{aligned}
\end{equation}
where $\mathbf{\Phi}_{\mathcal{I}}^{A}$ and $\mathbf{Z}_{\mathcal{I}}$
collect the student descriptors and projected teacher features indexed by $\mathcal{I}$, respectively.
The coefficient $\lambda_{\mathrm{mse}}$ balances the feature regression, 
and $\widetilde{\mathbf{x}}=\mathbf{x}/\lVert\mathbf{x}\rVert_{2}$ denotes $L_{2}$ normalization.

To directly penalize teacher-student angular deviation,
we additionally measure their geodesic angle on the unit hypersphere:
\begin{equation}
  \theta_i = \arccos \!\left(
    \langle \widetilde{\boldsymbol{\phi}}_i^{A}, \widetilde{\mathbf{z}}_i \rangle
    \right),\qquad i\in \mathcal{I}.
\end{equation}
The hyperspherical refinement loss is  
\begin{equation}
\label{eq:l_sph}
\mathcal{L}_{\text{sph}} = 
\frac{1}{|\mathcal{I}|} \sum_{i \in \mathcal{I}} \big[\theta_i + \max(0, \; \theta_i - m)^2\big],
\end{equation}
where $m$ is the angular threshold. 
This loss promotes pointwise angular alignment to the projected teacher
and emphasizes pairs with large angular discrepancy; it does not by itself preserve pairwise teacher angles
(Spearman correlations are reported in the Supplementary Material).

\paragraph{Rigid-View Descriptor Consistency}
The index-wise correspondences between $\mathcal{P}_{A}$ and $\mathcal{P}_{B}$ 
provide self-supervision for descriptors 
that remain consistent under rigid transformations
(occlusion and resampling robustness are addressed later in Stage~2).
We sample $M_{\mathrm{rv}}$ index-aligned points from the two views 
and denote their row-wise $L_{2}$-normalized descriptor matrices by
$\widetilde{\mathbf{\Phi}}_{\mathrm{rv}}^{A}, \widetilde{\mathbf{\Phi}}_{\mathrm{rv}}^{B} 
\in \mathbb{R}^{M_{\mathrm{rv}} \times d}$.
A symmetric InfoNCE objective~\cite{oord2018representation} encourages corresponding descriptors 
to remain similar while separating non-corresponding points:
\begin{equation}
\label{eq:l_infonce}
\begin{aligned}
\mathcal{L}_{\text{InfoNCE}} = \frac{1}{2}
\Bigg[
  &\operatorname{CE}\left(\frac{\widetilde{\mathbf{\Phi}}_{\mathrm{rv}}^{A}
  \left(\widetilde{\mathbf{\Phi}}_{\mathrm{rv}}^{B}\right)
  ^{\top}}{\tau},\mathbf{y}\right)\\
+ & \operatorname{CE} \left(\frac{\widetilde{\mathbf{\Phi}}_{\mathrm{rv}}^{B}
\left(\widetilde{\mathbf{\Phi}}_{\mathrm{rv}}^{A}\right)
^{\top}}{\tau},\mathbf{y}\right)
\Bigg],
\end{aligned}
\end{equation}
where $\tau$ is the temperature,
$\mathbf{y}=(0,1,\ldots,M_{\mathrm{rv}}-1)$ contains the correspondence indices,
and $\operatorname{CE}$ denotes the row-wise cross-entropy loss.

To further reduce descriptor drift, 
we introduce a soft rigid-transformation invariance objective 
over all $N$ index-aligned point pairs: 
\begin{equation}
\label{eq:l_eq}
\mathcal{L}_{\mathrm{eq}}=\frac{1}{N}\sum_{i=1}^{N}\left(1-
\left\langle
\widetilde{\boldsymbol{\phi}}_{i}^{A},
\widetilde{\boldsymbol{\phi}}_{i}^{B}
\right\rangle
\right).
\end{equation}
While $\mathcal{L}_{\mathrm{InfoNCE}}$ promotes point-wise discriminability 
through positive and negative pairs,
$\mathcal{L}_{\mathrm{eq}}$ directly enforces descriptor consistency
between corresponding points across rigidly transformed views.

\paragraph{Stage-1 Objective}

The complete Stage-1 objective combines cross-modal semantic transfer 
with self-supervised rigid-view consistency:
\begin{equation}
\label{eq:l_stage1}
\mathcal{L}_{\mathrm{stage1}} = w_d \mathcal{L}_{\mathrm{distill}} 
+ w_s \mathcal{L}_{\mathrm{sph}} 
+ w_n \mathcal{L}_{\mathrm{InfoNCE}} 
+ w_e \mathcal{L}_{\mathrm{eq}},
\end{equation}
where $w_d, w_s, w_n,$ and $w_e$ are non-negative balancing coefficients. 
The cross-modal losses are restricted to the active visibility indices $\mathcal{I}$, 
while the rigid-view losses operate on index-aligned point pairs
between $\mathcal{P}_{A}$ and $\mathcal{P}_{B}$, 
using $M_{\mathrm{rv}}$ sampled pairs for $\mathcal{L}_{\mathrm{InfoNCE}}$ 
and all $N$ pairs for $\mathcal{L}_{\mathrm{eq}}$.

\section{Stage-2: Pairwise Registration Fine-Tuning}
\label{subsec:stage2}

\begin{figure*}[t]
\centering
\includegraphics[width=\linewidth]{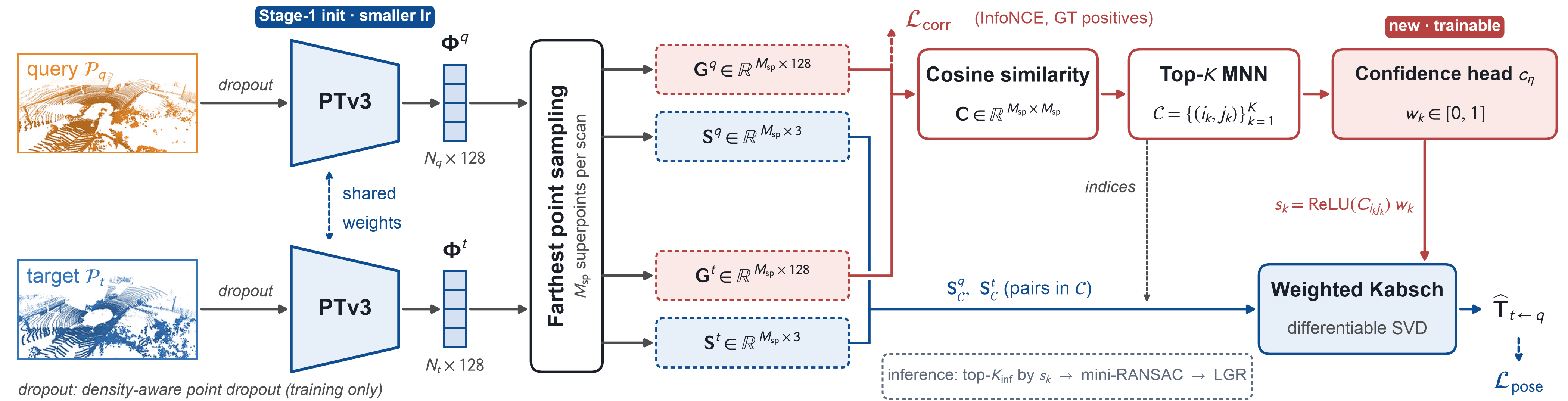}
\caption{Stage-2 registration fine-tuning.
Shared PTv3 descriptors, FPS superpoints, MNN matches with scalar confidence weights, and differentiable Kabsch in training;
inference uses mini-RANSAC then LGR.}
\label{fig:stage2}
\end{figure*}

Stage 2 adapts the distilled LiDAR representation to pairwise registration using only 3D inputs
(Figure~\ref{fig:stage2}).
Given an unposed query--target pair
$\mathcal{P}_{q}=
\{\mathbf{p}_{i}^{q}\}_{i=1}^{N_q}$
and
$\mathcal{P}_{t}
=
\{\mathbf{p}_{j}^{t}\}_{j=1}^{N_t}$,
a weight-shared Siamese PTv3 backbone, initialized from Stage 1,
extracts point-wise descriptors from both scans.
The vision teacher and image inputs are no longer used in this stage.
To preserve the distilled representation during registration-specific adaptation, 
the pretrained backbone is updated with a smaller learning rate 
than the newly initialized correspondence and confidence modules.
The network is jointly optimized through correspondence and pose supervision, 
together with density-aware point-dropout augmentation.

\paragraph{Pairwise Superpoint Generation}

Given the query and target scans, 
the Stage-1-pretrained, weight-shared PTv3 backbone extracts dense descriptor matrices
$\mathbf{\Phi}^{q}\in\mathbb{R}^{N_q\times d}$ and
$\mathbf{\Phi}^{t}\in\mathbb{R}^{N_t\times d}$, respectively.
To reduce the cost of pairwise matching, 
we apply Farthest Point Sampling (FPS) to the point coordinates 
and select $M_{\mathrm{sp}}$ representative points from each scan.
This yields the superpoint coordinates $\mathbf{S}^{q},\mathbf{S}^{t}\in\mathbb{R}^{M_{\mathrm{sp}}\times 3}$
and their associated descriptors $\mathbf{G}^{q},\mathbf{G}^{t}\in\mathbb{R}^{M_{\mathrm{sp}}\times d}$.
These superpoints serve as the basis for correspondence estimation
and pose recovery.

\paragraph{Confidence-Weighted Correspondence Estimation}

We select the top-$K$ mutual nearest-neighbor (MNN) correspondences 
based on the pairwise cosine similarity between the superpoint descriptors.
Let $\mathcal{C}=\{(i_k,j_k)\}_{k=1}^{K}$ denote the selected correspondence indices, 
and let $C_{i_k,j_k}$ denote the descriptor similarity of the $k$-th pair.
Let $\mathbf{g}_{i}^{q}$ and $\mathbf{g}_{j}^{t}$ denote the $i$-th and $j$-th rows of
$\mathbf{G}^{q}$ and $\mathbf{G}^{t}$, respectively.
Because high descriptor similarity does not necessarily indicate 
consistency with the underlying rigid transformation, 
a confidence head predicts an additional weight for each candidate correspondence:
\begin{equation}
    w_k
    =
    c_{\eta}
    \left(
    \left[
    \mathbf{W}_{c}\mathbf{g}_{i_k}^{q}
    \,\Vert\,
    \mathbf{W}_{c}\mathbf{g}_{j_k}^{t}
    \right]
    \right)
    \in [0,1],
\end{equation}
where $\mathbf{W}_{c}$ is a shared feature projection,
$c_{\eta}$ denotes the confidence head, and $\Vert$ denotes feature
concatenation. The final correspondence score is defined as
\begin{equation}
    s_k
    =
    \operatorname{ReLU}
    \left(
    C_{i_k,j_k}
    \right)
    \,
    w_k.
\end{equation}
The scores $\{s_k\}_{k=1}^{K}$ weight the corresponding 3D point pairs 
during differentiable pose estimation.

\paragraph{Density-Aware Point Dropout}

To expose the model to variations in scan sparsity, point dropout is
independently applied to the query and target scans during fine-tuning
with probability $p_{\mathrm{apply}}$.
When point dropout is applied to a scan containing $N$ points,
we sample a keep ratio
$\rho \sim \mathcal{U}[\rho_{\min},1]$
and retain a random subset of size
\begin{equation}
    N_{\mathrm{keep}}
    =
    \max
    \left(
    N_{\min},
    \left\lfloor \rho N \right\rfloor
    \right).
\end{equation}
This augmentation encourages the model to remain robust to variations 
in point density and sampling topology.
Implementation values ($p_{\mathrm{apply}}$, $\rho_{\min}$, $N_{\min}$) and a dropout-strength ablation are provided in the Supplementary Material.

\paragraph{End-to-End Registration Objective} 

During training, positive cross-scan superpoint pairs are identified by spatial proximity 
after transforming the query coordinates with the ground-truth relative pose.
The corresponding normalized descriptors are optimized 
using the same symmetric InfoNCE formulation as Equation~\eqref{eq:l_infonce}, 
while the remaining sampled descriptors serve as negatives.
We denote this cross-scan descriptor objective by $\mathcal{L}_{\mathrm{corr}}$.

For pose estimation, the coordinates associated with $\mathcal{C}$ are collected as
\begin{equation*}
    \mathbf{S}_{\mathcal{C}}^{q}
    =
    \left[
    \mathbf{S}_{i_k}^{q}
    \right]_{k=1}^{K},
    \qquad
    \mathbf{S}_{\mathcal{C}}^{t}
    =
    \left[
    \mathbf{S}_{j_k}^{t}
    \right]_{k=1}^{K}
\end{equation*}
and $\boldsymbol{\alpha}=(s_1,\ldots,s_K)^{\top}$ collects their correspondence scores.
A differentiable weighted Kabsch solver~\cite{kabsch1976solution} estimates the relative pose:
\begin{equation}
    \widehat{\mathbf{T}}_{t\leftarrow q}
    =
    \left(
    \widehat{\mathbf{R}},
    \widehat{\mathbf{t}}
    \right)
    =
    \operatorname{Kabsch}
    \left(
    \mathbf{S}_{\mathcal{C}}^{q},
    \mathbf{S}_{\mathcal{C}}^{t},
    \boldsymbol{\alpha}
    \right).
    \label{eq:weighted_kabsch}
\end{equation}
Given the ground-truth relative pose
$\mathbf{T}_{t\leftarrow q}^{\mathrm{gt}}=(\mathbf{R}_{\mathrm{gt}},\mathbf{t}_{\mathrm{gt}})$,
the pose loss is defined as
\begin{equation}
\begin{aligned}
    \mathcal{L}_{\text{pose}}
    ={}&
    \arccos
    \left(
    \frac{
    \operatorname{tr}
    \left(
    \widehat{\mathbf{R}}^{\top}
    \mathbf{R}_{\mathrm{gt}}
    \right)
    -1
    }{2}
    \right)
    +
    \lambda_{t}
    \left\lVert
    \widehat{\mathbf{t}}
    -
    \mathbf{t}_{\mathrm{gt}}
    \right\rVert_{2},
\end{aligned}
\end{equation}
where $\lambda_t$ balances the rotational and translational errors.
The rotation term measures geodesic distance on $\mathrm{SO}(3)$, 
while the translation term uses Euclidean distance.

The complete Stage-2 objective is
\begin{equation}
    \mathcal{L}_{\mathrm{stage2}}
    =
    \mathcal{L}_{\mathrm{corr}}
    +
    \lambda_{p}
    \mathcal{L}_{\mathrm{pose}},
    \label{eq:l_stage2}
\end{equation}
where $\lambda_p$ balances descriptor-level correspondence learning and pose-level supervision.
Because weighted Kabsch is differentiable with respect to the selected correspondence scores, 
gradients from $\mathcal{L}_{\mathrm{pose}}$ update the confidence head and LiDAR descriptor backbone 
through the continuous weighting path, without requiring binary inlier labels.

\section{Inference}
\label{sec:inference}

At inference, CVSD-Reg uses only LiDAR inputs.
Following the Stage-2 pipeline, we extract
$M_{\mathrm{inf}}=2048$ FPS superpoints from each scan and retain the
$K_{\mathrm{inf}}=512$ correspondences with the highest scores $s_k$.
A vectorized three-point RANSAC generates
$N_{\mathrm{r}}=20{,}000$ pose hypotheses and selects the transformation
with the largest inlier support under a distance threshold
$\delta_{\mathrm{r}}=1.5\,\mathrm{m}$ as the initial estimate $\mathbf{T}^{(0)}$.
Starting from $\mathbf{T}^{(0)}$, 
we refine the pose using the Local Geometric Refinement (LGR) procedure of GeoTransformer~\cite{geotransformer}.
At each iteration, correspondences whose residuals under the current transformation 
exceed $\delta_{\mathrm{lgr}}$ are assigned zero weight,
while the remaining pairs retain their correspondence scores.
The pose is then re-estimated using weighted Kabsch.
We perform $N_{\mathrm{lgr}}=50$ iterations with
$\delta_{\mathrm{lgr}}=0.6\,\mathrm{m}$.
The same inference configuration is used for all benchmarks without
camera inputs, external pose solvers, or post-hoc ICP refinement.

\section{Experiments}
\label{sec:exp}

\subsection{Experimental Setup}
\label{subsec:exp-setup}

\paragraph{Training protocol.}
Stage~1: nuScenes train (6 cameras)$+$KITTI 00--07. Stage~2: KITTI 00--07 / nuScenes train (eval: KITTI 08--10 / nuScenes val).
HeLiPR is unused for training. Learned baselines use official checkpoints (KITTI-trained where applicable; BUFFER-X released; none saw nuScenes);
HeLiPR comparisons are zero-shot for all methods. Details are in the Supplementary Material.

\paragraph{Evaluation metrics.}
The primary evaluation metric is strict Success Rate (SR@0.5\,m/$1^\circ$),
complemented by relaxed thresholds (SR@1\,m/$2^\circ$ and SR@2\,m/$5^\circ$).
A scan pair is counted as a successful registration if its relative translation error (RTE) and relative rotation error (RRE) both fall below the respective threshold.
We additionally report median RTE/RRE over all evaluated pairs without conditioning on registration success,
using synchronized GNSS/INS pose ground truth as the reference.
Consequently, a method can post a low median RTE while recording $0\%$ success rate if many failures are driven by rotation rather than translation.
\subsection{Evaluation Benchmarks}
\label{subsec:benchmarks}

\paragraph{KITTI Odometry} We evaluate performance on the standard test sequences (08, 09,
and 10) of the KITTI odometry benchmark~\cite{kitti}, comprising 555 evaluation pairs.
Scans are captured using a 64-beam Velodyne HDL-64E sensor under high-overlap conditions.
This benchmark serves to measure the baseline registration accuracy in a standard in-distribution setting.

\paragraph{nuScenes} We sample 500 key-frame pairs with a trajectory gap of 1--5\,m from the validation split of the nuScenes dataset~\cite{nuscenes}.
This benchmark evaluates registration performance on 32-beam spinning LiDAR scans.

\paragraph{HeLiPR Cross-Sensor} To evaluate zero-shot cross-sensor generalization,
we use held-out HeLiPR~\cite{helipr} sequences with Ouster-128 reference maps and queries from four LiDAR types
(Velodyne-16, Livox Avia, Aeva FMCW, and Ouster-128), totaling $600$ pairs ($150$ per sensor).
Pair construction details are provided in the Supplementary Material.

\begin{table*}[t]
\centering
\caption{%
\textbf{Zero-shot cross-sensor generalization on HeLiPR.}
}
\label{tab:helipr_crosssensor}
\resizebox{\textwidth}{!}{%
\setlength{\tabcolsep}{4pt}
\begin{tabular}{l cccc c cccc cc}
\toprule
\multirow{3}{*}{Method}
& \multicolumn{4}{c}{Primary: SR@0.5m/$1^\circ$ ($\%$) $\uparrow$} & \phantom{x} & \multicolumn{4}{c}{Relaxed: SR@1m/$2^\circ$ ($\%$) $\uparrow$}
& \multicolumn{2}{c}{Median Error} \\
\cmidrule(lr){2-5}\cmidrule(lr){7-10}\cmidrule(lr){11-12}
& Ouster & Velodyne & Avia & Aeva &
& Ouster & Velodyne & Avia & Aeva
& RTE (m)$\downarrow$ & RRE ($^\circ$)$\downarrow$ \\
& (Same) & 16-beam & (Solid) & (FMCW) &
& (Same) & 16-beam & (Solid) & (FMCW) & & \\
\midrule
\shortstack[l]{FPFH + TEASER \\ \cite{rusu2009fast,yang2020teaser}} & 100.0 &   2.7 &  30.0 &  60.0 &&  100.0 &  16.0 &  53.3 &  86.7 & 1.080 & 3.42 \\
KISS-Matcher~\cite{lim2025kiss}                   & 100.0 &   1.3 &   6.0 &  31.3 &&  100.0 &  45.3 &  36.0 &  70.0 & 0.515 & 1.95 \\
RAP~\cite{rap}                                    &  94.0 &  12.7 &  11.3 &   8.7 &&  100.0 &  27.3 &  46.0 &  44.0 & 1.061 & 3.28 \\
CAST~\cite{cast}                                  & 100.0 &   0.0 &   0.0 &   0.0 &&  100.0 &   0.0 &   0.0 &   0.0 & 6.115 & 48.6 \\
GeoTransformer~\cite{geotransformer}              & 100.0 &  60.0 &   0.0 &   0.0 &&  100.0 &  80.0 &   0.0 &   0.0 & 0.164 & 11.2 \\
UGP~\cite{ugp}                                    & 100.0 &  52.7 &  18.7 &  45.3 &&  100.0 &  78.7 &  37.3 &  56.0 & 0.470 & 1.68 \\
PARE-Net~\cite{parenet}                           & 100.0 &  15.3 &  37.3 &  27.3 &&  100.0 &  44.0 &  61.3 &  54.7 & 0.535 & 1.91 \\
BUFFER-X~\cite{seo2025buffer}                     & 100.0 &  53.3 &  82.7 &  94.0 &&  100.0 &  91.3 & 100.0 &  99.3 & 0.269 & 0.74 \\
\midrule
\rowcolor[gray]{0.95} \textbf{CVSD-Reg (Ours)}
& \textbf{100.0} & \textbf{97.3} & \textbf{100.0} & \textbf{100.0} &&
\textbf{100.0} & \textbf{98.0} & \textbf{100.0} & \textbf{100.0} & \textbf{0.107} & \textbf{0.218} \\
\bottomrule
\end{tabular}%
}
\end{table*}

\begin{table}[t]
\centering
\caption{%
\textbf{Registration performance on the KITTI odometry benchmark}
}
\label{tab:kitti_full}
\resizebox{\columnwidth}{!}{%
\setlength{\tabcolsep}{4pt}
\begin{tabular}{l ccc cc}
\toprule
\multirow{2}{*}{Method}
& \multicolumn{3}{c}{Success Rate ($\%$)}
& \multicolumn{2}{c}{Median Error} \\
\cmidrule(lr){2-4}\cmidrule(lr){5-6}
& SR@2m/5$^\circ$ & SR@1m/2$^\circ$ & SR@0.5m/1$^\circ$
& RTE (m) $\downarrow$ & RRE ($^\circ$) $\downarrow$ \\
\midrule
FPFH + TEASER  & 98.7 & 97.7 & 92.3 & 0.070 & 0.295 \\
MAC                            & 93.0 & 88.8 & 78.4 & 0.095 & 0.410 \\
RAP                             & 97.1 & 93.5 & 81.8 & 0.263 & 0.366 \\
KISS-Matcher                    & 97.1 & 87.7 & 66.3 & 0.140 & 0.723 \\
UGP                            & 99.8 & 99.6 & 97.7 & 0.060 & 0.189 \\
PARE-Net                    & 99.8 & 99.6 & 97.7 & 0.040 & 0.172 \\
GeoTransformer        & 99.8 & 99.6 & 97.7 & 0.054 & 0.173 \\
BUFFER-X                     & 99.8 & 99.6 & 97.1 & 0.081 & 0.207 \\
CAST                            & \textbf{100.0} & \textbf{100.0} & \textbf{99.3} & \textbf{0.023} & \textbf{0.125} \\
\midrule
\rowcolor[gray]{0.95} \textbf{CVSD-Reg (Ours)} & 98.7 & 98.7 & 97.7 & 0.047 & 0.153 \\
\bottomrule
\end{tabular}%
}
\end{table}

\begin{table}[t]
\centering
\caption{%
\textbf{Registration performance on nuScenes benchmark}
}
\label{tab:nuscenes_full}
\resizebox{\columnwidth}{!}{%
\setlength{\tabcolsep}{4.5pt}
\begin{tabular}{l ccc cc}
\toprule
\multirow{2}{*}{Method}
& \multicolumn{3}{c}{Success Rate ($\%$)}
& \multicolumn{2}{c}{Median Error} \\
\cmidrule(lr){2-4}\cmidrule(lr){5-6}
& SR@2m/5$^\circ$ & SR@1m/2$^\circ$ & SR@0.5m/1$^\circ$
& RTE (m) $\downarrow$ & RRE ($^\circ$) $\downarrow$ \\
\midrule
FPFH + TEASER  & 87.4 & 87.4 & 85.6 & 0.090 & 0.302 \\
KISS-Matcher                    & 78.4 & 77.2 & 67.8 & 0.124 & 0.523 \\
GeoTransformer       & 99.4 & 98.6 & 97.0 & 0.080 & \textbf{0.239} \\
BUFFER-X                     & 92.6 & 90.6 & 89.6 & 0.113 & 0.294 \\
\midrule
\rowcolor[gray]{0.95} \textbf{CVSD-Reg (Ours)} & \textbf{100.0} & \textbf{100.0} & \textbf{99.0} & \textbf{0.067} & 0.249 \\
\bottomrule
\end{tabular}%
}
\end{table}

\subsection{Main Results}
\label{sec:exp-main}

\begin{figure}[t]
\centering
\includegraphics[width=\columnwidth]{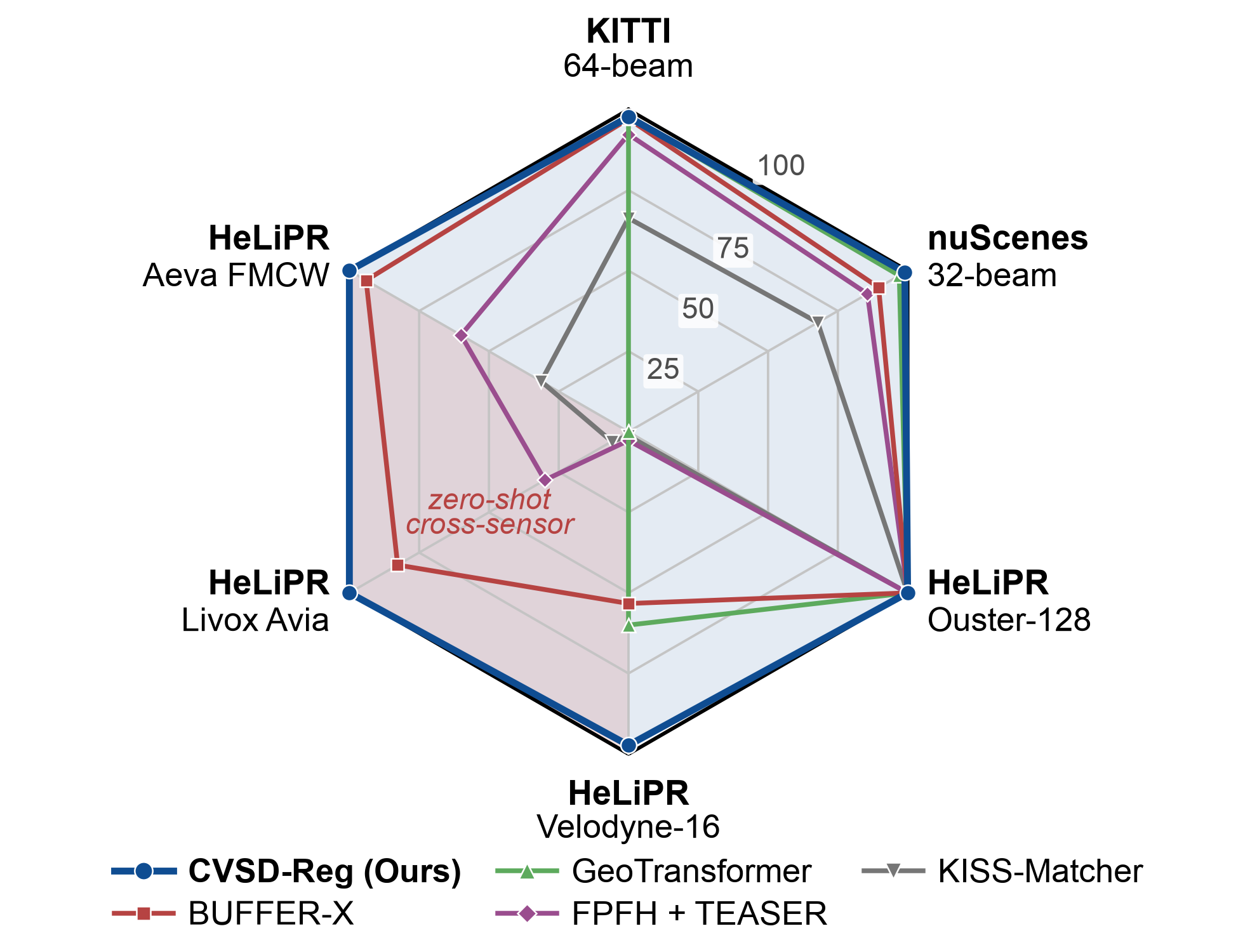}
\caption{%
\textbf{Strict success rate (SR@0.5\,m/$1^\circ$) across all benchmarks and HeLiPR sensors.}
CVSD-Reg (blue, filled) nearly saturates every axis with a single checkpoint and inference recipe, whereas purely geometric methods collapse on the cross-sensor axes (Velodyne-16, Avia, Aeva). Only methods reported across Tables~\ref{tab:helipr_crosssensor},~\ref{tab:kitti_full}, and~\ref{tab:nuscenes_full} are shown.}
\label{fig:radar}
\end{figure}

Tables~\ref{tab:helipr_crosssensor},~\ref{tab:kitti_full}, and~\ref{tab:nuscenes_full} and Figure~\ref{fig:radar} summarize registration success rates and median pose errors on HeLiPR, KITTI, and nuScenes under a single CVSD-Reg checkpoint.
We highlight two takeaways, namely that in-distribution competence is retained while zero-shot cross-sensor robustness improves substantially over geometric baselines.

\paragraph{Cross-Sensor Robustness}
Table~\ref{tab:helipr_crosssensor} isolates zero-shot registration of queries from four LiDAR types against an Ouster reference map.
Relative to BUFFER-X~\cite{seo2025buffer}, a strong overall geometric baseline on HeLiPR, CVSD-Reg gains $+44.0\,\text{pp}$ on sparse Velodyne-16 ($97.3\%$ vs.\ $53.3\%$; $+37.3\,\text{pp}$ over GeoTransformer~\cite{geotransformer}, the best baseline on that sensor),
$+17.3\,\text{pp}$ on solid-state Avia ($100.0\%$ vs.\ $82.7\%$),
and $+6.0\,\text{pp}$ on FMCW Aeva ($100.0\%$ vs.\ $94.0\%$),
for $+16.8\,\text{pp}$ overall ($99.3\%$ vs.\ $82.5\%$).
Median RTE/RRE also improve from $0.269$\,m / $0.74^\circ$ to $0.107$\,m / $0.218^\circ$.
The largest gap appears precisely where purely geometric cues are sparsest or least regular, such as in 16-beam scans and non-repetitive patterns,
which is consistent with semantic priors disambiguating matches when local geometry alone is unreliable.
The same checkpoint transfers zero-shot to TIERS (VLP-16/Avia/Horizon queries against an OS0-128 map; $95.3\%$ vs.\ $71.0\%$ for BUFFER-X) and runs in $0.42$\,s per pair on one RTX~4090; details are in the Supplementary Material.


\paragraph{Universal vs.\ Specialist Capabilities}
Tables~\ref{tab:kitti_full} and~\ref{tab:nuscenes_full} show that geometric specialists remain strong when test geometry matches training.
On KITTI, GeoTransformer~\cite{geotransformer}, UGP~\cite{ugp}, and PARE-Net~\cite{parenet} all reach $97.7\%$ strict success rate, with CAST~\cite{cast} reaching the peak at $99.3\%$ ($0.023$\,m / $0.125^\circ$ median error).
However, these same specialists collapse under HeLiPR sensor shifts, as illustrated in Table~\ref{tab:helipr_crosssensor} and Figure~\ref{fig:radar}; for instance, GeoTransformer and CAST both drop to $0\%$ on Avia and Aeva.
This highlights a clear accuracy--generality trade-off that CVSD-Reg successfully addresses.
CVSD-Reg matches the competitive $97.7\%$ KITTI benchmark at $0.047$\,m / $0.153^\circ$ median error,
leads nuScenes with $99.0\%$ strict success rate compared to GeoTransformer ($97.0\%$) and BUFFER-X~\cite{seo2025buffer} ($89.6\%$),
and remains near-saturated on HeLiPR with $97.3\%$ on Velodyne and $99.3\%$ overall.
This supports our claim that distilling visual semantic priors improves cross-sensor generalization without sacrificing competitive in-distribution accuracy.

\subsection{Ablation Studies}
\label{sec:exp-ablation}

\begin{table}[t]
\centering
\caption{%
\textbf{Stage-1 distillation ablation on HeLiPR}
(strict success rate at $0.5$\,m and $1^\circ$. Overall is the mean over sensors).
}
\label{tab:distill_helipr}
\resizebox{\columnwidth}{!}{%
\setlength{\tabcolsep}{4pt}
\begin{tabular}{lccccc}
\toprule
Method & Overall & Ouster & Vel 16 & Avia & Aeva \\
\midrule
Stage-2 only (no distill, from scratch) & 16.2 & 58.0 & 6.7 & 0.0 & 0.0 \\
LiDAR-only Stage~1 (no teacher) & 41.5 & 98.0 & 32.7 & 14.0 & 21.3 \\
\midrule
\rowcolor[gray]{0.95} \textbf{CVSD-Reg (Ours)}
& \textbf{99.3} & \textbf{100.0} & \textbf{97.3} & \textbf{100.0} & \textbf{100.0} \\
\bottomrule
\end{tabular}%
}
\end{table}

\paragraph{Necessity of Semantic Distillation}
Table~\ref{tab:distill_helipr} compares CVSD-Reg with two matched controls that share the Stage-2 recipe, data, dropout, and inference solver.
Stage-2-only training from scratch reaches $16.2\%$ overall on held-out HeLiPR.
A LiDAR-only Stage~1 control (same PTv3, KITTI$+$nuScenes, $100$k steps, rigid-view augmentation, and $\mathcal{L}_{\text{InfoNCE}}+\mathcal{L}_{\text{eq}}$, but no DINOv2 teacher) reaches $41.5\%$ ($32.7\%$ on Velodyne-16).
CVSD-Reg reaches $99.3\%$, so the progression $16.2\%\rightarrow 41.5\%\rightarrow 99.3\%$ isolates most of the cross-sensor gain to visual semantic supervision rather than pretraining alone.
Descriptor-level statistics before the pose solver and a TEASER$+$LGR variant are reported in the Supplementary Material.
\begin{table}[t]
\centering
\caption{%
\textbf{Test-time DINOv2 projection vs.\ distilled descriptors}
(strict success rate at $0.5$\,m and $1^\circ$ on a $100$-pair KITTI diagnostic subset).
}
\label{tab:vfm}
\resizebox{\columnwidth}{!}{%
\setlength{\tabcolsep}{6pt}
\begin{tabular}{l cc}
\toprule
\textbf{Descriptor Source} & \textbf{Strict success rate ($\%$) $\uparrow$} & \textbf{Camera-Free?} \\
\midrule
DINOv2-L projection (test-time) & 24.0 & No \\
\rowcolor[gray]{0.95} \textbf{CVSD-Reg (Ours)} & \textbf{98.0} & \textbf{Yes} \\
\bottomrule
\end{tabular}%
}
\end{table}

\paragraph{Cross-Modal Distillation vs.\ Test-Time VFM Projection}
Table~\ref{tab:vfm} compares raw test-time DINOv2 projection against our distilled LiDAR student on a fixed $100$-pair KITTI diagnostic subset (not the $555$-pair protocol of Table~\ref{tab:kitti_full}).
Both rows use the same lightweight matching and optimization pipeline so that only the descriptor source differs. Absolute success rates are therefore not comparable to the main KITTI numbers.
Direct projection requires cameras at test time and yields only $24.0\%$ strict success rate, whereas CVSD-Reg reaches $98.0\%$ camera-free.
This supports that structured distillation, rather than na\"ive feature lifting, produces correspondence-ready LiDAR descriptors.

\begin{table}[!tb]
\centering
\caption{Stage-1 loss ablation on a $210$-pair HeLiPR subset (Velodyne/Avia/Aeva).
Overall is the mean over sensors. Full denotes the complete Stage-1 objective, including $\mathcal{L}_{\text{InfoNCE}}$.}
\label{tab:abl}
\resizebox{\columnwidth}{!}{%
\begin{tabular}{lcccc}
\toprule
Configuration & Overall & Vel 16 & Avia & Aeva \\
\midrule
$\mathcal{L}_{\text{distill}}$  & 79.5 & 68.6 & 84.3 & 85.7 \\
\,\,$+\mathcal{L}_{\text{eq}}$         & 83.3 & 75.7 & 87.1 & 87.1 \\
\,\,$+\mathcal{L}_{\text{sph}}$        & 85.7 & 81.4 & 88.6 & 87.1 \\
Raise cosine wt.\ (no $\mathcal{L}_{\text{sph}}$) & 80.5 & 70.0 & 85.7 & 85.7 \\
\rowcolor[gray]{0.95} \textbf{Full CVSD-Reg (Ours)} & \textbf{88.1} & \textbf{84.3} & \textbf{91.4} & \textbf{88.6} \\
\bottomrule
\end{tabular}%
}
\end{table}

\paragraph{Stage-1 Loss Formulations} We ablate Stage-1 loss terms on a $210$-pair heterogeneous subset of HeLiPR ($70$ pairs per sensor for Velodyne, Avia, and Aeva),
with results summarized in Table~\ref{tab:abl}.
Each entry is $k/70$ successes to one decimal place, and Overall is the mean over the three sensors.
Unlike Table~\ref{tab:helipr_crosssensor}, this diagnostic excludes Ouster and uses a smaller pair set,
so absolute rates are not comparable to the main HeLiPR protocol (e.g., Full CVSD-Reg $88.1\%$ here vs.\ $99.3\%$ there).
Relative to distill-only, $\mathcal{L}_{\text{sph}}$ mainly helps Velodyne-16 ($+12.8\,\text{pp}$) and $\mathcal{L}_{\text{eq}}$ adds overall stability ($+3.8\,\text{pp}$).
Raising the cosine weight without $\mathcal{L}_{\text{sph}}$ reaches only $80.5\%$ vs.\ $85.7\%$ for $+\mathcal{L}_{\text{sph}}$, so the margin term is not redundant.
Full Stage~1 attains the best overall rate ($+8.6\,\text{pp}$); a ScaLR-style control ($78.1\%$) is in the Supplementary Material.

\section{Conclusion}
\label{sec:conclusion}

We presented CVSD-Reg, which distills visual semantic priors into LiDAR descriptors for robust global registration.
Decoupling Stage-1 semantic pretraining from Stage-2 registration adaptation yields a camera-free pipeline that remains accurate under density, sampling, and sensor shifts.
Matched controls attribute most cross-sensor gains to the visual teacher; a remaining limitation is calibrated image--LiDAR data.

\bibliography{refs}
\clearpage
\appendix
\setcounter{table}{0}
\setcounter{figure}{0}
\renewcommand{\thetable}{A\arabic{table}}
\renewcommand{\thefigure}{A\arabic{figure}}
\newcommand{\appsection}[1]{\refstepcounter{section}\section*{\thesection\quad #1}}


\appsection{Training and Evaluation Details}
\label{app:training}

\paragraph{Stage-1 pretraining.}
We pretrain on the nuScenes train split (paired RGB--LiDAR, 6 cameras) and KITTI sequences 00--07 for 100k steps using AdamW (learning rate $10^{-4}$, cosine decay) with the final layer of DINOv2 (ViT-L/14) as the distillation target.
The Stage-1 loss weights are $w_d = 1.0$, $w_s = 0.3$, $w_n = 0.5$, and $w_e = 0.3$ (for $\mathcal{L}_{\text{distill}}, \mathcal{L}_{\text{sph}}, \mathcal{L}_{\text{InfoNCE}}, \mathcal{L}_{\text{eq}}$, respectively).
Images are resized to $448 \times 896$; DINOv2 outputs a $1024 \times 32 \times 64$ feature map (per-point teacher features $\mathbf{f}_i \in \mathbb{R}^{1024}$ after bilinear sampling and multi-view averaging).
For visual feature lifting, a LiDAR point is treated as visible in a camera if its depth lies in $[z_{\min}, z_{\max}]$\,m with $z_{\min}=0.5$ and $z_{\max}=120$, and the projected pixel falls inside the image bounds.
The PTv3 student voxelizes inputs at $0.1$\,m and projects decoder features to $128$-d descriptors.
Rigid-view augmentation samples $\mathbf{T}_{\mathrm{aug}}$ with a rotation about a random axis of angle uniform in $[-15^\circ, 15^\circ]$ and translations uniform in $[-2,2]^{3}$\,m.
For $\mathcal{L}_{\text{InfoNCE}}$, we sample $M_{\mathrm{rv}}{=}2048$ index-aligned anchors with temperature $\tau{=}0.07$.

\paragraph{Stage-2 fine-tuning.}
We fine-tune for 20k steps on KITTI (sequences 00--07) and nuScenes training-scene pairs with per-epoch data-source balancing.
KITTI evaluation uses sequences 08--10; nuScenes evaluation uses validation-scene pairs only.
We apply a differential learning rate: trunk multiplier $0.05$ ($2.5{\times}10^{-6}$) and head learning rate $5{\times}10^{-5}$.
Point-dropout augmentation is applied independently to each scan with probability $p_{\mathrm{apply}}{=}0.5$:
we sample $\rho\sim\mathcal{U}[\rho_{\min},1]$ with $\rho_{\min}{=}0.25$ and keep
$N_{\mathrm{keep}}=\max(N_{\min},\lfloor\rho N\rfloor)$ points ($N_{\min}{=}500$).
$\mathcal{L}_{\text{pose}}$ is computed using the top-$128$ correspondences by score.
HeLiPR is excluded from all training stages.

\paragraph{Evaluation protocols.}
\textbf{KITTI:} 555 consecutive-frame pairs from sequences 08--10 (64-beam Velodyne HDL-64E).
\textbf{nuScenes:} 500 validation key-frame pairs with a 1--5\,m gap (32-beam LiDAR).
\textbf{HeLiPR:} held-out KAIST05, DCC05, and RIVER05; Ouster-128 reference maps with Velodyne VLP-16, Livox Avia, Aeva FMCW, and in-distribution Ouster queries ($150$ pairs per sensor; $600$ pairs total).
All methods share identical pair lists under unified $\text{xyz}+\text{intensity}$ inputs and no ICP.
``Standardized'' denotes a shared channel layout, not per-sensor intensity normalization: raw intensity is used as-is, mixing KITTI $[0,1]$ and nuScenes $[0,255]$ scales in training.
Zeroing intensity at test time yields $98.8\%$ HeLiPR SR ($96.0\%$ on Velodyne-16) vs.\ $99.3\%$, so robustness does not rely on intensity matching.
Learned baselines use official checkpoints (KITTI-trained where applicable; BUFFER-X released model; none saw nuScenes).
The primary metric is strict SR@0.5\,m/$1^\circ$, complemented by relaxed thresholds and median RTE/RRE against GNSS/INS ground truth.

\paragraph{Hyperparameter summary.}
Table~\ref{tab:hparams} lists the settings used for all reported CVSD-Reg numbers.
A single checkpoint and a single inference recipe are shared across KITTI, nuScenes, and HeLiPR.

\begin{table}[t]
\centering
\caption{\textbf{CVSD-Reg implementation details.} One checkpoint and one inference recipe are used for KITTI, nuScenes, and HeLiPR.}
\label{tab:hparams}
\resizebox{\columnwidth}{!}{%
\setlength{\tabcolsep}{5pt}
\begin{tabular}{@{}ll@{}}
\toprule
\textbf{Component} & \textbf{Setting} \\
\midrule
\multicolumn{2}{l}{\textit{Backbone (PTv3 student)}} \\
Input channels            & 4 (xyz $+$ intensity) \\
Voxel size                & $0.1$\,m \\
Stem                      & sparse conv $4 \to 32$ \\
Encoder channels          & $(32, 64, 128, 256, 512)$ \\
Encoder depths            & $(2, 2, 2, 6, 2)$ \\
Serialization             & Z-order $+$ Hilbert (and transposed) \\
Descriptor dim $d$        & $128$ \\
\midrule
\multicolumn{2}{l}{\textit{Teacher (frozen)}} \\
Model                     & DINOv2 ViT-L/14 (\texttt{facebook/dinov2-large}) \\
Image resolution          & $448 \times 896$ \\
Feature map               & $1024 \times 32 \times 64$ \\
Projector                 & bias-free $1024 \to 128$ \\
\midrule
\multicolumn{2}{l}{\textit{Stage-1 pretraining}} \\
Data                      & nuScenes train $+$ KITTI 00--07 (paired RGB--LiDAR) \\
Steps / optimizer         & 100k / AdamW, lr $10^{-4}$, cosine \\
Loss weights              & $w_d{=}1.0,\ w_s{=}0.3,\ w_n{=}0.5,\ w_e{=}0.3$ \\
Spherical margin $m$      & $0.2$\,rad \quad InfoNCE $\tau$ $=0.07$ \\
InfoNCE anchors $M_{\mathrm{rv}}$ & $2048$ \\
Rigid-view aug            & random-axis rot.\ $\in[-15^\circ,15^\circ]$, $t\sim\mathcal{U}[-2,2]^{3}$\,m \\
$\lambda_{\text{mse}}$    & $0.1$ \\
\midrule
\multicolumn{2}{l}{\textit{Stage-2 fine-tuning}} \\
Data                      & KITTI (00--07) $+$ nuScenes train, balanced \\
Steps                     & 20k \\
LR (trunk / head)         & $2.5{\times}10^{-6}$ (mult.\ $0.05$) / $5{\times}10^{-5}$ \\
Superpoints (FPS)         & $M_{\mathrm{sp}} = 1024$ (training) \\
Corr.\ top-$K$ (train)    & $1024$ \\
Loss                      & $\mathcal{L}_{\text{corr}} + \lambda_{p}\mathcal{L}_{\text{pose}}$, $\lambda_{p}{=}0.5$ \\
Pose loss                 & geodesic rot.\ $+\ \lambda_{t}\|{\Delta t}\|_2,\ \lambda_{t}{=}0.1$ \\
Pose top-$K$              & 128 correspondences by score \\
Point dropout             & $p_{\mathrm{apply}}{=}0.5,\ \rho_{\min}{=}0.25,\ N_{\min}{=}500$ \\
\midrule
\multicolumn{2}{l}{\textit{Inference (unified recipe)}} \\
Superpoints (FPS)         & $M_{\mathrm{inf}} = 2048$ \\
Correspondences           & top-$K_{\mathrm{inf}} = 512$ by score \\
mini-RANSAC               & 3-point, $N_{\mathrm{r}}{=}20{,}000$, $\delta_{\mathrm{r}}{=}1.5$\,m \\
Refinement                & LGR, $N_{\mathrm{lgr}}{=}50$, $\delta_{\mathrm{lgr}}{=}0.6$\,m \\
Post-hoc ICP              & none \\
\bottomrule
\end{tabular}%
}
\end{table}

\appsection{Additional Controls and Analyses}
\label{app:controls}

\paragraph{ScaLR-style distillation.}
On the same $210$-pair HeLiPR subset as the main-paper Stage-1 ablation, a faithful ScaLR-style recipe
(same PTv3 student, DINOv2 teacher, data, budget, Stage~2, and solver; student projected to the teacher dimension; cosine alignment without MSE)
reaches $78.1\%$ overall ($65.7\%$ on Velodyne-16), versus $88.1\%$ / $84.3\%$ for full Stage~1
($+10.0$ / $+18.6$\,pp).

\paragraph{Pairwise geometry retention.}
Although $\mathcal{L}_{\text{sph}}$ is pointwise, Spearman correlation of pairwise cosines on a Stage-1 student is $0.91$ with the projected teacher and $0.74$ with raw DINOv2.

\paragraph{Descriptor quality vs.\ pose solver.}
Under the matched mini-RANSAC$+$LGR recipe of the main paper, Stage-2-only vs.\ CVSD-Reg is $16.2\%$ vs.\ $99.3\%$ on HeLiPR.
With TEASER$+$LGR instead (Table~\ref{tab:app-dropout}), CVSD-Reg still reaches $98.7\%$ overall.
Before any pose solver, the top-$512$ inlier ratio is $46.8\%$ for CVSD-Reg vs.\ $7.9\%$ for Stage-2-only,
and a single weighted Kabsch step (no RANSAC/LGR) already yields $81.3\%$ strict SR for CVSD-Reg.

\paragraph{Runtime.}
Per pair on one RTX~4090, CVSD-Reg takes $0.42$\,s with $N_r{=}20{,}000$ RANSAC
(backbone $0.09$, RANSAC $0.24$, LGR $0.09$), vs.\ $0.31$\,s for GeoTransformer and $1.85$\,s for BUFFER-X.
Reducing RANSAC to $5$k hypotheses yields $0.24$\,s total at $98.5\%$ HeLiPR SR (vs.\ $99.3\%$ at $20$k).

\paragraph{Harder HeLiPR pairs and TIERS.}
On HeLiPR pairs with $5$--$10$\,m query--map offsets, CVSD-Reg retains $94.7\%$ SR vs.\ $61.3\%$ for BUFFER-X
(median nearest-map gap on the standard protocol is $0.8$\,m with $72.4\%$ overlap on Velodyne/Avia/Aeva).
The same checkpoint evaluated zero-shot on TIERS (VLP-16/Avia/Horizon vs.\ OS0-128, $300$ pairs)
reaches $95.3\%$ vs.\ $71.0\%$ for BUFFER-X.

\appsection{Point-Dropout Ablation}
\label{app:ablation}

Table~\ref{tab:app-dropout} isolates Stage-2 dropout strength under a fixed TEASER$+$LGR solver,
so that differences reflect the training augmentation rather than the pose initializer used in the main paper.
For readability we parameterize strength by $d_{\max}=1-\rho_{\min}$ (larger $d_{\max}$ allows stronger downsampling).
Activating dropout at $\rho_{\min}{=}0.25$ ($d_{\max}{=}0.75$) improves Velodyne strict SR from $91.3\%$ to $97.3\%$ (LGR 50).
The main-paper HeLiPR numbers instead use the unified mini-RANSAC$+$LGR recipe on the same dropout-trained checkpoint.

\begin{table}[t]
\centering
\caption{Point-dropout strength ablation on HeLiPR (600 pairs; TEASER$+$LGR).
Overall is the mean of Ouster/Velodyne/Avia/Aeva (Ouster is $100.0\%$ in all rows and omitted from the columns).
$d_{\max}=1-\rho_{\min}$ follows the main-paper keep-ratio parameterization.
Main-paper HeLiPR results use mini-RANSAC instead.}
\label{tab:app-dropout}
\resizebox{\columnwidth}{!}{%
\setlength{\tabcolsep}{3.5pt}
\begin{tabular}{@{}cc cccc c@{}}
\toprule
$d_{\max}$ & \shortstack{LGR\\iters} & Overall & Vel. & Avia & Aeva & \shortstack{Med.\\RTE (m) $\downarrow$} \\
& & \multicolumn{4}{c}{SR@0.5m/$1^\circ$ ($\%$) $\uparrow$} & \\
\midrule
\multirow{2}{*}{0 (none)}
& 20 & 96.3 & 91.3 & 96.0 & 98.0 & 0.121 \\
& 50 & 96.8 & 91.3 & 97.3 & 98.7 & 0.118 \\
\midrule
\multirow{2}{*}{0.50}
& 20 & 98.2 & 95.3 & 97.3 & \textbf{100.0} & \textbf{0.114} \\
& 50 & 98.5 & 96.7 & 97.3 & \textbf{100.0} & \textbf{0.114} \\
\midrule
\multirow{2}{*}{0.75}
& 20 & 98.2 & 96.7 & 96.7 & 99.3 & 0.117 \\
& 50 & \textbf{98.7} & \textbf{97.3} & \textbf{98.0} & 99.3 & 0.118 \\
\bottomrule
\end{tabular}%
}
\end{table}

\appsection{Qualitative Cross-Sensor Registration}
\label{app:qualitative}

\begin{figure*}[t]
\centering
\includegraphics[width=\textwidth]{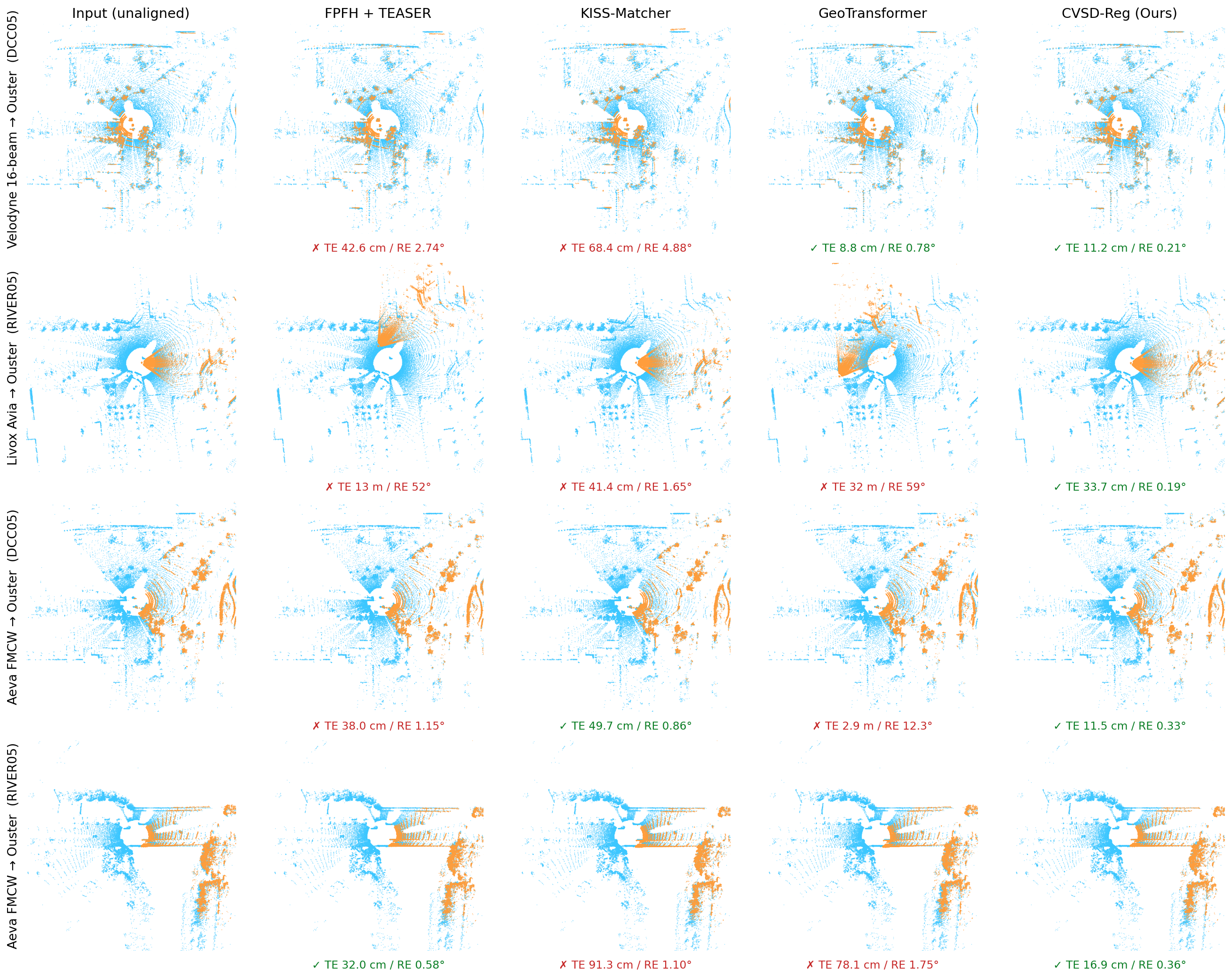}
\caption{%
\textbf{Qualitative HeLiPR cross-sensor registration.}
Each row shows a different query$\rightarrow$map pairing against an Ouster-128 reference (blue); the transformed query is orange.
Columns compare unaligned input, FPFH$+$TEASER, KISS-Matcher, GeoTransformer, and \textbf{CVSD-Reg (Ours)}.
Per-cell RTE/RRE report translation and rotation errors; \checkmark/\textbf{\texttimes} indicate success at SR@0.5\,m/$1^\circ$.
}
\label{fig:app-qual-helipr}
\end{figure*}

Figure~\ref{fig:app-qual-helipr} visualizes representative HeLiPR cross-sensor pairs.
Geometric matchers degrade on non-repetitive Livox Avia and FMCW Aeva patterns, whereas CVSD-Reg aligns structure across all four sensor pairings.

\appsection{HeLiPR Pair Construction}
\label{app:helipr}

We evaluate on the held-out KAIST05, DCC05, and RIVER05 sequences of HeLiPR~\cite{helipr}.
Dense 128-beam Ouster scans serve as the reference map,
while query streams cover four sensor architectures:
a sparse 16-beam Velodyne VLP-16,
a non-repetitive solid-state Livox Avia,
an FMCW Aeva,
and an in-distribution Ouster-128.
Query frames are spatio-temporally matched to the nearest Ouster reference map frame using synchronized GNSS/INS,
subject to a spatial gate of $<5$\,m.
This yields exactly 50 pairs per sequence per sensor
($3$ sequences $\times$ $50$ queries $\times$ $4$ sensors $= 600$ pairs; $150$ pairs per sensor).
All clouds are standardized to $\text{xyz}+\text{intensity}$ before evaluation,
and the same pair lists are shared across all compared methods.

\end{document}